\documentclass[letterpaper]{article} % DO NOT CHANGE THIS
\usepackage{aaai25}  % DO NOT CHANGE THIS
\usepackage{times}  % DO NOT CHANGE THIS
\usepackage{helvet}  % DO NOT CHANGE THIS
\usepackage{courier}  % DO NOT CHANGE THIS
\usepackage[hyphens]{url}  % DO NOT CHANGE THIS
\usepackage{graphicx} % DO NOT CHANGE THIS
\usepackage{natbib}  % DO NOT CHANGE THIS AND DO NOT ADD ANY OPTIONS TO IT
\usepackage{caption} % DO NOT CHANGE THIS AND DO NOT ADD ANY OPTIONS TO IT
\usepackage{algorithm}
\usepackage{algorithmic}

\usepackage{xspace}
\makeatletter
\DeclareRobustCommand\onedot{\futurelet\@let@token\@onedot}
\def\@onedot{\ifx\@let@token.\else.\null\fi\xspace}
 
\def\ie{\emph{i.e}\onedot}

\makeatother

\usepackage[T1]{fontenc}

\usepackage{booktabs}
\usepackage{multirow}

\usepackage{amsfonts,amssymb, amsmath}

\usepackage{newfloat}
\usepackage{listings}
\DeclareCaptionStyle{ruled}{labelfont=normalfont,labelsep=colon,strut=off} % DO NOT CHANGE THIS
\floatstyle{ruled}
\newfloat{listing}{tb}{lst}{}
\floatname{listing}{Listing}
\title{Transfer Learning of Real Image Features with Soft Contrastive Loss \\for Fake Image Detection}
\author{
    Ziyou Liang\textsuperscript{\rm 1}, 
    Weifeng Liu\textsuperscript{\rm 1},
    Run Wang\textsuperscript{\rm 1}\thanks{Corresponding author. E-mail: wangrun@whu.edu.cn}, 
    Mengjie Wu\textsuperscript{\rm 1}, 
    Boheng Li\textsuperscript{\rm 2},\\ 
    Yuyang Zhang\textsuperscript{\rm 1}, 
    Lina Wang\textsuperscript{\rm 1}, 
    Xinyi Yang\textsuperscript{\rm 3}
}
\affiliations{
    \textsuperscript{\rm 1}Key Laboratory of Aerospace Information Security and Trusted Computing, Ministry of Education, \\School of Cyber Science and Engineering, Wuhan University, China\\
    \textsuperscript{\rm 2}Nanyang Technological University, Singapore\\
    \textsuperscript{\rm 3}NSFOCUS, China
    
}

\usepackage{bibentry}
\begin{document}

\maketitle

\begin{abstract}
In the last few years, the artifact patterns in fake images synthesized by different generative models have been inconsistent, leading to the failure of previous research that relied on spotting subtle differences between real and fake. In our preliminary experiments, we find that the artifacts in fake images always change with the development of the generative model, while natural images exhibit stable statistical properties. In this paper, we employ natural traces shared only by real images as an additional target for a classifier. Specifically, we introduce a self-supervised feature mapping process for natural trace extraction and develop a transfer learning based on soft contrastive loss to bring them closer to real images and further away from fake ones. This motivates the detector to make decisions based on the proximity of images to the natural traces. To conduct a comprehensive experiment, we built a high-quality and diverse dataset that includes generative models comprising GANs and diffusion models, to evaluate the effectiveness in generalizing unknown forgery techniques and robustness in surviving different transformations. Experimental results show that our proposed method gives \textbf{96.2\%} mAP significantly outperforms the baselines. Extensive experiments conducted on popular commercial platforms reveal that our proposed method achieves an accuracy exceeding \textbf{78.4\%}, underscoring its practicality for real-world application deployment. 
% The source code and partial self-built dataset are available in supplementary material.
\end{abstract}


\section{Introduction}

With the rapid development and maturity of generative models, the increasing proliferation of fake images has attracted widespread attention. Compared to Generative Adversarial Networks (GANs), diffusion models (DMs), as today's SOTA generative models, exhibit better generation quality~\cite{ADM} and even support powerful text-to-image models such as DALL-E2~\cite{DALLE2}, Stable Diffusion~\cite{LDM}. Currently, one can use different types of generative models to create realistic faces or complex scene images and it is foreseeable that more generative models for image synthesis will emerge in the future. Therefore, it is the goal for the community to develop a more practical method to distinguish fake images that are synthesized with unknown forgery techniques as the unseen generative models will emerge inadvertently.

\begin{figure}[t]
\centering
\includegraphics[width=0.95\linewidth]{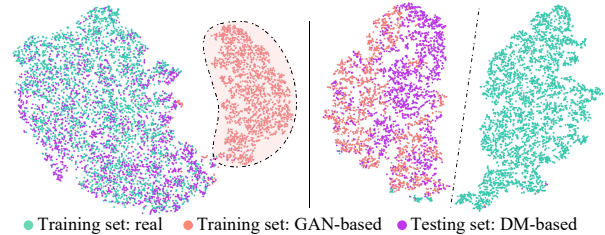} 
% \vspace{-5pt}
\caption{\small\textbf{TSNE visualization.} \textbf{Left}: The detector is easily overfitted to fake images in the training set; \textbf{Right}: Training with natural traces can generalize to unknown fake images.}
% \vspace{-15pt}
\label{fig: fake trace}
\end{figure}

The prior paradigm for fake image detection is to learn artifacts of fake images by capturing the subtle differences between real and fake images ~\cite{wang2019fakespotter, CNNDetection, huang2020fakepolisher, wang2021faketagger}. Our empirical research confirms that there is a huge challenge in detecting unknown fake images: the model can easily form a classification manifold on the images of the training set, while new generative fake images and their distinctive artifacts will be randomly scattered in this space, as shown in Figure \ref{fig: fake trace} left. Unfortunately, the existing studies are trapped in the endless efforts to spot the artifacts by investigating the subtle differences between real and fake \cite{corvi2023detection}. Since the classifier can easily overfit the fake image artifacts of the training set, unknown fake images will be scattered outside the manifold like real images.
% In order to solve this generalization problem, current research can be divided into 1) continuous learning and 2) transfer learning, where transfer learning is to apply the knowledge or patterns learned in a certain field to different but related fields or problems. Common transfer learning methods include feature-based and parameter-based. In recent years, methods based on pre-training have also been widely used. As far as we know, there is still a gap in the exploration of transfer learning in fake image detection.

% Figure: 现有方法是利用统一伪影实现检测的

In this paper, we propose a transfer learning method based on natural trace features for fake image detection. We propose a novel perspective on fake image detection, emphasizing \textit{the inherent similarities among real images rather than the differences between real and fake}. Our insight is grounded in the premise that images from the real world possess a stable intrinsic naturalness. We posit that real images share common characteristics, akin to those found in fake images, which we refer to as "natural traces." We introduce natural trace forensics (NTF) and adopt a substitution strategy to replace shared features with homogeneous features. Specifically, we exploit a self-supervised feature mapping process to decouple heterogeneous features of real images, allowing for the capture of stable homogeneous features. We then develop a transfer learning for homogeneous features based on a soft contrastive loss, which simultaneously solved the problem that similar clusters formed by self-supervised contrastive learning are still scattered with large distances and outliers under the constraints of supervised contrastive learning do not converge. We jointly optimized transfer training and binary classification to force the classifier to aggregate homogeneous features close to real images and constrain them away from fake images. In Figure \ref{fig: fake trace} right, the classifier reduces its dependence on specific artifact patterns and gains the ability to detect images synthesized by unknown generative models with natural traces.

To better evaluate whether our proposed method can be well generalized to other unknown generative models, we build a generated image dataset consisting of 12 SOTA generative models, including 6 GANs and 6 DMs. Our dataset covers a variety of categories, such as faces and scenes, to evaluate detectors' capability across various types of fake images. In addition, we also evaluate the performance of our method in identifying images generated by Midjourney\footnote{https://www.midjourney.com/} and Kolors\footnote{https://klingai.kuaishou.com/}, which is currently a popular commercial tool for text-to-image generation. Experimental results show that our proposed method could discriminate GAN-based, DM-based, and Multi-step fake images (synthesized with at least two different generated models) in high confidence with an average accuracy of more than \textbf{96.1\%} and is sufficiently robust to various image perturbation transformations.

Our main contributions are summarized as follows:

\begin{itemize}
\item We are the first to propose soft contrastive transfer learning, which utilizes the disentangled shared features of real images for fake image detection. Our results also demonstrate its effectiveness in detecting previously unseen generative fake images, highlighting its potential as a versatile solution in this domain.
\item To conduct a comprehensive evaluation, we build a dataset including GAN-based, DM-based, and Multi-step manipulation for generating fake images. For the first time, we generate Multi-step fake images by employing multiple synthesis methods. 
% This self-built dataset will be released to the community to advance the community to develop more effective and practical methods for distinguishing unseen forgery techniques.
\item Experimental results show the effectiveness of our proposed method in tackling fake images generated by SOTA GANs and diffusion models, giving an average precision of more than 96.18\%, significantly outperforming the baselines. 
% Additionally, our method is robust against the common image transformations and identifies the undistinguishable Midjourney in high confidence.
\end{itemize}
% \vspace{-5pt}

\section{Related Work}

\subsection{Transfer Learning in Fake Detection}

In recent years, the most common method of transfer learning is to fine-tune the pre-trained model on the target dataset. \cite{suratkar2023deep} detect fake videos through the utilization of transfer learning in autoencoders and a hybrid model of CNN and RNN. \cite{lee2022fake, ghayoomi2022deep} used pre-trained BiLSTM\cite{zhou2016attention} and RoBerta\cite{liu2019roberta} for transfer learning on fake news about COVID-19 in Korean and Persian, respectively.
Although the fine-tuning method is highly flexible for downstream tasks, it is also prone to catastrophic forgetting. Freezing layers of the pre-trained backbone network and finetuning only lateral fully connected layers is one of the most effective transfer learning methods. \cite{ranjan2020improve} explored the effectiveness and interpretability of freezing convolutional layers and fine-tuning fully connected layers in DeepFake detection, and \cite{elhassan2022dft} studied the transfer of lip-motion-based DeepFake detection methods on 11 models. However, the pre-training of these frozen transfer learning methods is obviously isolated from the target, and the parameter transfer effect is limited. As far as we know, there is still a gap in the exploration of transfer learning in fake image detection. Our method balances the feature transfer and classification by additional soft contrastive loss constraints.

% \subsection{Fake Images Synthesis}

% In recent years, generative models represented by GANs and diffusion models have been evolving towards maturity. GANs train a generator and a discriminator to compete with each other until the generator can produce realistic images, while DMs utilize diffusion and denoising Markov chains for training, surpassing GANs in training stability, diversity, and quality of generation~\cite{maze2023diffusion,stypulkowski2024diffused}. This marks an evolution from producing or manipulating facial appearances, commonly termed DeepFakes~\cite{juefei2022countering}, to generating contextually relevant text or image prompts. What's more, with the emergence of new generative models, the artifacts of fake images are varied and more imperceptible. It is foreseeable that the vast majority of artifact-based detection will fail when the fake image artifacts are unseen. This progression has escalated public apprehension regarding the challenges in identifying fake images. In this paper, we are dedicated to developing a practical method for detecting fake images that are manipulated by unknown generative models.

\subsection{Fake Images Detection}

Existing research focuses on exploring subtle differences between real and fake images, and these can be categorized into explicit and implicit artifact-based methods.

\noindent{\textbf{Explicit-based.}} Some researchers noticed that they often contained specific artifacts or unnatural patterns\cite{AutoGAN, liu2023towards}. Additionally, several studies focused on exploring GAN-based artifacts in the frequency domain\cite{frank2020leveraging} and the failure to accurately re-enact certain biological features when generating fake faces\cite{hu2021exposing,tan2023deepfake}. These findings have motivated researchers to use explicit artifacts to detect fake images through simple classifiers. However, as generative models are continuously updated and iterated, these artifacts become imperceptible or even disappear, making detection methods reliant on explicit artifacts less effective.

\noindent{\textbf{Implicit-based.}} Wang \textit{et al.}\cite{CNNDetection} demonstrated that with appropriate training data and data augmentation, neural networks could detect other GAN-based images, while \cite{Tan_2023_CVPR} used CNNs to transform images into gradient form to present a broader range of artifacts. These studies indicate that images generated by both GAN and diffusion models possess distinct ``fingerprints" different from the real images\cite{Yu_2019_ICCV,sha2023fake}. Nevertheless, there is still insufficient evidence to prove that generative models from different families have universal fingerprints for detection.

% In this work, we propose an innovative method: training with real images for fake detection. It's noteworthy that ~\cite{liu2022detecting} has similarities with ours, where they detect fake by analyzing noise patterns in real images. However, their method relies on the noise patterns of various devices. Our method uses natural traces extracted from real images for training. Compared to noise patterns, which may vary with different devices, can be complex to analyze, and are susceptible to intentional modification for evasion, natural traces consistently present in real images offer greater robustness.

% table: dataset
\begin{table}[t]
\scriptsize
\begin{center}
% \vspace{-5pt}
\renewcommand{\arraystretch}{0.8}
\setlength{\tabcolsep}{1pt}
\begin{tabular*}{\hsize}{cccccc}
\toprule
\textbf{Family} & \textbf{Type} & \textbf{Method} & \textbf{Year} & \textbf{Image Source} & \textbf{\# Images} \\ \midrule
% \hdashline
& \multirow{5}{*}{Unconditional} & ProGAN & 2017 & CelebA-HQ & 4.0k \\
& & StyleGAN2 & 2019 & CelebA-HQ/FFHQ/LSUN & 12.0k \\
GAN- & & ProjGAN & 2021 & FFHQ/LSUN/Landscape & 12.0k \\
based & & VQGAN & 2020 & CelebA-HQ & 4.0k \\
& & Diff-StyleGAN2 & 2022 & FFHQ/LSUN & 8.0k \\
\cmidrule(lr){2-6}
& Image-to-Image & SimSwap & 2020 & CelebA-HQ & 4.0k \\ \midrule
% \hdashline
& \multirow{3}{*}{Unconditional} & DDPM & 2020 & CelebA-HQ/LSUN & 8.0k \\
& & DDIM & 2021 & CelebA-HQ & 4.0k \\
DM- & & PNDM & 2022 & CelebA-HQ/LSUN & 12.0k \\
\cmidrule(lr){2-6}
based & Image-to-Image & DiffFace & 2022 & CelebA-HQ & 2.0k \\
\cmidrule(lr){2-6}
& \multirow{2}{*}{Prompt-guided} & LDM  & 2022 & CelebA-HQ/LAION & 8.0k \\
& & SDM & 2022 & LAION & 4.0k \\ \midrule
% \hdashline
& \multirow{2}{*}{GAN-GAN} & SimSwap\_Style2 & 2024 & CelebA-HQ & 2.0k \\
& &  SimSwap\_VQ & 2024 & CelebA-HQ & 2.0k \\
\cmidrule(lr){2-6}
Multi- & \multirow{3}{*}{GAN-DM} &  SimSwap\_LDM & 2024 & CelebA-HQ & 2.0k \\
step & &  DiffFace\_Style2 & 2024 & CelebA-HQ & 2.0k \\
& &  DiffFace\_Proj & 2024 & FFHQ & 2.0k \\
\cmidrule(lr){2-6}
& DM-DM &  DiffFace\_LDM & 2024 & CelebA-HQ & 2.0k \\
\bottomrule
\end{tabular*}
% \vspace{-10pt}
\caption{\small Statistics of the self-built dataset, including GAN-based, DM-based, and Multi-step synthesis.}
% \vspace{-15pt}
\label{tab: dataset}
\end{center}
\end{table}

\section{A Diverse Generated Image Dataset}

Due to the lack of DM-based generated images in current datasets, we create a dataset with a wide range of generative models. This dataset aims to enhance the evaluation of fake detection methods' capability. It includes fake images generated by various models, alongside an equal number of real images from corresponding training sets for each method.

Our dataset covers two major families: GANs and diffusion models, with each fake image synthesized using one generative model. Particularly, we have developed a novel multi-step fake image generation method, involving collaboration between two or more generative models, to achieve identity swapping or attribute editing between real and fake faces. To ensure diversity, the dataset comprises various categories of generation methods: unconditional generation, image-to-image generation, and prompt-guided generation, as shown in Table \ref{tab: dataset}. Moreover, our carefully selected generative models exhibit fundamental differences in their generations, ensuring extensive representation of the dataset.

\noindent\textbf{GAN-based forgery}: We select six representative GANs to generate forgery images including unconditional and image-to-image. The innovation of ProGAN~\cite{Progan} in introducing progressive training has a significant impact on subsequent research, while StyleGAN2~\cite{Styleganv2} refines style control through decoupling image features and generates more realistic images. In terms of architectural improvements, ProjGAN~\cite{Projectedgan} enhances generator feedback using pre-trained weights, while VQGAN~\cite{Vqgan} and Diff-StyleGAN2~\cite{Diffusiongan} innovatively replace the backbone network with transformer and diffusion processes, respectively, resulting in higher image quality and more stable training processes. SimSwap's (ss)\cite{chen2020simswap} ID injection module achieves breakthroughs in arbitrary face swapping. 

\noindent\textbf{DM-based forgery}: DM-based surpasses GAN-based in terms of image quality and diversity. Here, we select six different DMs for creating fake images including unconditional, image-to-image, and prompt-guided. DDPM\cite{DDPM}, as the initial diffusion model, lays the groundwork, with DDIM\cite{DDIM} and PNDM\cite{PNDM} enhancing execution speed and quality. DiffFace (df)\cite{kim2022diffface} is the first identity-conditioned DDPM that uses diffusion models for face swapping. And LDM\cite{LDM}, which employs pre-trained self-encoders to map pixels to latent space, combined with context learning in cross-attention layers for prompt-guided image generation, along with Stable Diffusion (SDM), a popular LDM-based prompt-guided model, represent further advancements.

\noindent\textbf{Multi-step}: In the real scenario, the creator tends to employ multiple forgery techniques to achieve better forgery. We create a Multi-step face synthesis dataset where two face-swapping methods (SimSwap and DiffFace) based on GAN or diffusion models swap real faces onto synthetic ones, including three hybrid modes: GAN-GAN, GAN-DM, and DM-DM. This generation method provides simulated threats of artifact disappearance or blending in real-world scenarios.

% In summary, our dataset includes high-resolution images ranging from $256\times256$ to $1024\times1024$ resolution. Furthermore, diverse image categories support a more comprehensive evaluation of universal fake detection.
% The dataset will be released to advance this interesting research field.

% Figure: method
\begin{figure*}[t]
\centering
\includegraphics[width=0.9\linewidth]{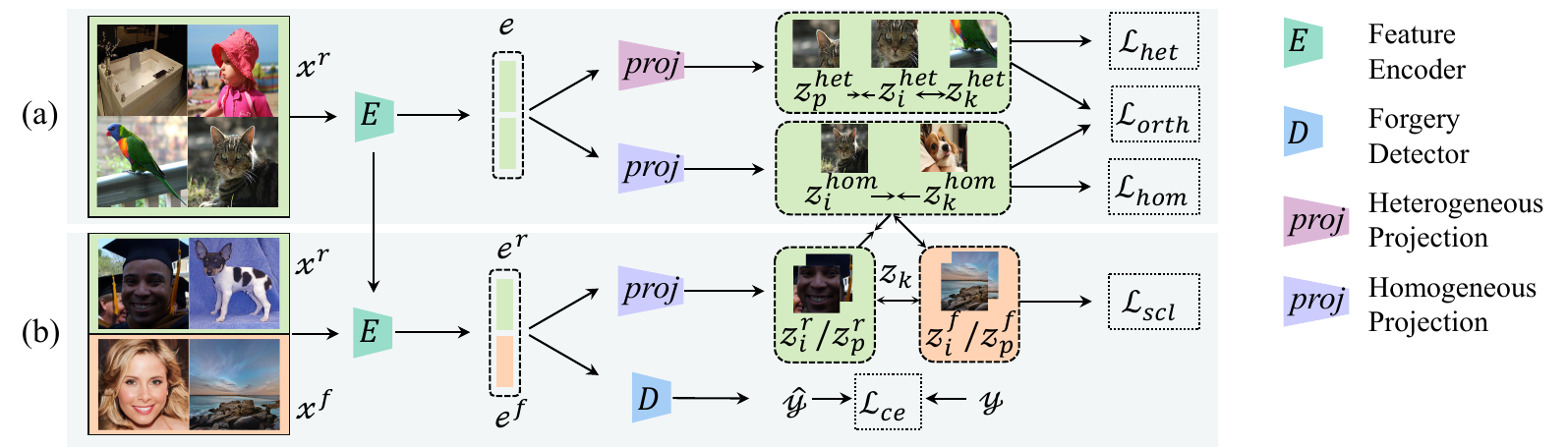}
% \vspace{-5pt}
\caption{\small\textbf{NTF architecture}. We first decouple the feature representation of real images into homogeneous and heterogeneous features. Next, the homogeneous features $z^{hom}$ participate in SCL with the real and fake image features. Detector classifies real and fake images guided by the target of intra-class aggregation and inter-class separation. Better view in color.}
\label{fig: method}
% \vspace{-10pt}
\end{figure*}

\begin{table*}[t]
\scriptsize
\centering
% The values in black show the cross-generator generalization. Of these, the highest values are highlighted in black.
\setlength{\tabcolsep}{0.7pt}
\begin{tabular}{cccccccccccccccccccccccccccccc}
\toprule
\multirow{2}{*}{\textbf{Category}} & \multirow{2}{*}{\textbf{Method}} & \multicolumn{4}{c}{\textbf{ProGAN}} & \multicolumn{4}{c}{\textbf{StyleGAN2}} & \multicolumn{4}{c}{\textbf{VQGAN}} & \multicolumn{4}{c}{\textbf{ProjGAN}} & \multicolumn{4}{c}{\textbf{Diff-StyleGAN2}} & \multicolumn{4}{c}{\textbf{SimSwap}} & \multicolumn{4}{c}{\textbf{Average}} \\ 
\cmidrule(lr){3-6}\cmidrule(lr){7-10}\cmidrule(lr){11-14}\cmidrule(lr){15-18}\cmidrule(lr){19-22}\cmidrule(lr){23-26}\cmidrule(lr){27-30}
& & AP $\uparrow$ & ACC $\uparrow$ & FPR $\downarrow$ & FNR $\downarrow$ & AP & ACC & FPR & FNR & AP & ACC & FPR & FNR & AP & ACC & FPR & FNR & AP & ACC & FPR & FNR & AP & ACC & FPR & FNR & AP & ACC & FPR & FNR \\
\multirow{2}{*}{Explicit-based} & SBIs* & 68.6 & 66.0 & 39.8 & 28.2 & 65.5 & 60.2 & 41.8 & 37.9 & 56.9 & 46.2 & 76.9 & 30.8 & 59.9 & 50.0 & 53.9 & 46.2 & 58.5 & 54.9 & 58.3 & 32.0 & 55.3 & 53.9 & \textbf{33.1} & 59.1 & 60.8 & 55.2 & 41.8 & 41.9\\
& CADDM* & 51.5 & 53.4 & 40.6 & 52.6 & 52.3 & 54.4 & 46.8 & 44.3 & 52.6 & 54.5 & 50.7 & 40.3 & 54.3 & 57.6 & 43.0 & 41.7 & 51.3 & 52.9 & 44.2 & 50.0 & 46.6 & 36.6 & 82.9 & 43.6 & 52.1 & 53.3 & 48.4 & 45.0 \\
\cmidrule(lr){1-30}
\multirow{5}{*}{Implicit-based} & Xception & \underline{99.3} & 84.5 & \textbf{0} & 31.1 & \textbf{100} & \underline{99.9} & \underline{0.1} & \textbf{0} & \textbf{100} & \textbf{99.9} & \underline{0.2} & \textbf{0} & \textbf{100} & \textbf{99.1} & \underline{1.8} & \textbf{0} & 97.9 & 84.0 & \underline{0.2} & 31.8 & \textbf{99.9} & \underline{90.1} & \underline{19.7} & \textbf{0} & \underline{99.5} & \textbf{92.9} & \underline{3.7} & 10.5 \\
& Wang2020 & 97.9 & 85.3 & 27.1 & 2.0 & 86.2 & 56.3 & 87.3 & \underline{0.1} & 61.2 & 50.5 & 98.9  & \underline{0.1} & 62.2 & 50.4 & 99.2 & \underline{0.1} & 87.8 & 68.5 & 61.3 & \underline{1.8} & 98.4 & 65.0 & 69.8 & \underline{0.1} & 82.3 & 62.7 & 73.9 & \underline{0.7} \\
& Grag2021 & \textbf{100} & \textbf{99.9} & \textbf{0} & \textbf{0.1} & \textbf{100} & \textbf{100} & \textbf{0} & \textbf{0} & \underline{99.9} & 91.6 & 16.8 & \underline{0.1} & 99.7 & \underline{94.6} & 10.8 & \textbf{0} & \underline{99.8} & \textbf{94.5} & 11.1 & \textbf{0} & 94.1 & 50.5 & 98.9 & \textbf{0} & 98.9 & 88.5 & 22.9 & \textbf{0}\\
& Ojha2023 & 98.8 & \underline{99.1} & \underline{1.5}  & \underline{0.5} & 83.1 & 87.0 & 15.9 & 10.1 & 83.6 & 87.3 & 16.1 & 9.4 & 75.2 & 78.7 & 33.2 & 9.4 & 64.6 & 64.9 & 69.3 & 0.8 & 80.3 & 83.8 & 23.5 & 8.9 & 80.9 & 83.5 & 26.6 & 6.5\\
\cmidrule(lr){2-30}
& \textit{\textbf{NTF(Ours)}} & \textbf{100} & 92.5 & \textbf{0} & 14.9 & \textbf{100} & 92.5 & \textbf{0} & 15.0 & \textbf{100} & \underline{92.6} & \textbf{0} & 14.9 & \underline{99.8} & 92.2 & \textbf{0.7} & 15.1 & \textbf{99.9} & \underline{92.7} & \textbf{0.2} & 14.6 & \underline{99.8} & \textbf{92.2} & \textbf{0.6} & 15.1 & \textbf{99.9} & \underline{92.4} & \textbf{0.2} & 14.9\\
\bottomrule
\end{tabular}
% \vspace{-10pt}
\caption{\small\textbf{Intra-family generalization on GANs}. Performance of NFT and baselines in spotting 6 GAN-based generated images. These baselines include the classic approach, XcetionNet, Wang2020, and Grag2021, as well as the latest methods, Ojha2023,  and CADDM. The method with $*$ means that the baseline is only evaluated on non-face images. The \textit{Average} column represents the weighted average of the corresponding metrics. Among those, the best and second-best performances are highlighted in \textbf{bold} and \underline{underlined}, respectively.}
\label{tab: GANbased}
% \vspace{-10pt}
\end{table*}

\begin{table*}[t]
\scriptsize
\centering
\setlength{\tabcolsep}{0.7pt}
\begin{tabular}{cccccccccccccccccccccccccccccc}
\toprule
\multirow{2}{*}{\textbf{Category}} & \multirow{2}{*}{\textbf{Method}} & \multicolumn{4}{c}{\textbf{DDPM}} & \multicolumn{4}{c}{\textbf{DDIM}} & \multicolumn{4}{c}{\textbf{PNDM}} & \multicolumn{4}{c}{\textbf{LDM}} & \multicolumn{4}{c}{\textbf{SDM}} & \multicolumn{4}{c}{\textbf{DiffFace}} & \multicolumn{4}{c}{\textbf{Average}} \\
\cmidrule(lr){3-6}\cmidrule(lr){7-10}\cmidrule(lr){11-14}\cmidrule(lr){15-18}\cmidrule(lr){19-22}\cmidrule(lr){23-26}\cmidrule(lr){27-30}
& & AP $\uparrow$ & ACC $\uparrow$ & FPR $\downarrow$ & FNR $\downarrow$ & AP & ACC & FPR & FNR & AP & ACC & FPR & FNR & AP & ACC & FPR & FNR & AP & ACC & FPR & FNR & AP & ACC & FPR & FNR & AP & ACC & FPR & FNR \\
\multirow{2}{*}{Explicit-based} & SBIs* & 63.1 & 64.0 & \underline{37.5} & 34.5 & 71.1 & 69.5 & \underline{15.5} & 47.6 & 64.2 & 59.2 & \underline{48.5} & 33.0 & \underline{83.8} & 75.7 & 14.6 & 34.0 & 57.0 & 55.8 & 59.2 & 29.1 & 70.6 & 56.9 & 28.0 & 28.4 & 68.3 & \underline{63.5} & 33.9 & 34.4 \\
& CADDM* & 53.0 & 55.2 & 46.5 & 43.2 & 55.8 & 59.8 & 35.8 & 44.6 & 51.8 & 52.5 & 51.9 & 43.1 & 52.3 & 53.9 & 48.3 & 43.9 & 51.6 & 54.5 & 38.3 & 52.5 & 53.0 & 52.6 & 43.9 & 44.2 & 52.9 & 54.7 & 44.1 & 45.2 \\
\cmidrule(lr){1-30}
\multirow{5}{*}{Implicit-based} & Xception & 74.7 & \textbf{72.3} & \textbf{5.1} & 50.2 & 75.8 & 73.9 & \textbf{12.4} & 39.9 & \textbf{89.6} & 49.8 & 99.9 & 0.6 & 65.8 & 65.0 & 8.8 & 61.2 & 41.1 & 40.3 & \underline{28.5} & 91.0 & 59.2 & 40.8 & 99.5 & 19.0 & 66.2 & 57.0 & \textbf{14.0} & 59.0\\
& Wang2020 & 58.2 & 50.2 & 99.7 & \underline{0.1} & 60.9 & 50.2 & 99.6 & \underline{0.1} & 66.4 & 50.0 & 99.6 & \underline{0.1} & 67.9 & 50.4 & 99.2 & \textbf{0} & 39.8 & 49.5 & 99.6 & \underline{1.5} & 34.8 & 49.8 & 99.8 & 2.4 & 54.6 & 50.0 & 99.6 & \underline{0.7}\\
& Grag2021 & \textbf{79.9} & 50.1 & 99.8 & \textbf{0} & \underline{83.7} & 50.3 & 99.5 & \textbf{0.1} & \underline{86.5} & 50.1 & 99.3 & \textbf{0} & \textbf{99.9} & \textbf{98.8} & \underline{2.5} & \textbf{0} & \underline{98.2} & \underline{63.6} & 72.9 & \textbf{0} & 63.2 & 54.4 & 91.3 & \textbf{0} & \underline{85.2} & 61.2 & 77.5 & \textbf{0}\\
& Ojha2023 & 64.2 & 67.0 & 57.4 & 8.6 & 71.8 & \underline{75.3} & 39.7 & 9.8 & 61.7 & \underline{64.4} & 62.2 & 9.0 & 80.2 & 83.9 & 22.9 & \underline{9.4} & 55.6 & 56.0 & 87.3 & 0.8 & \underline{88.4} & \textbf{88.7} & \underline{22.0} & \underline{0.6} & 70.3 & 72.6 & 48.6 & 6.3\\
\cmidrule(lr){1-30}
& \textit{\textbf{NTF(Ours)}} & \underline{76.9} & \underline{68.8} & 47.2 & 15.3 & \textbf{91.8} & \textbf{84.1} & 16.4 & \textbf{15.5} & 79.4 & \textbf{71.0} & \textbf{42.7} & 15.1 & \textbf{99.9} & \underline{92.7} & \textbf{0.1} & 14.7 & \textbf{99.5} & \textbf{92.1} & \textbf{1.1} & 14.8 & \textbf{99.3} & \underline{87.4} & \textbf{0.6} & 15.5 & \textbf{91.2} & \textbf{82.7} & \underline{18.0} & 15.1\\
\bottomrule
\end{tabular}
% \vspace{-10pt}
\caption{\small\textbf{Cross-family generalization on diffusion models}. Performance of NFT and baselines in spotting 6 DM-based generated images.}
\label{tab: DMbased}
% \vspace{-10pt}
\end{table*}
%Please note that \textit{Midjourney} exclusively gathers generated images, hence, only ACC and FPR scores are utilized for evaluation. The scores from \textit{Midjourney} are disregarded in the calculation of the \textit{Average}.

\begin{table*}[t]
\scriptsize
\centering
\setlength{\tabcolsep}{0.7pt}
\begin{tabular}{cccccccccccccccccccccccccccccc}
\toprule
\multirow{2}{*}{\textbf{Category}} & \multirow{2}{*}{\textbf{Method}} & \multicolumn{4}{c}{\textbf{SimSwap\_Style2}} & \multicolumn{4}{c}{\textbf{SimSwap\_VQ}} & \multicolumn{4}{c}{\textbf{SimSwap\_LDM}}  & \multicolumn{4}{c}{\textbf{DiffFace\_Style2}} & \multicolumn{4}{c}{\textbf{DiffFace\_Proj}}  & \multicolumn{4}{c}{\textbf{DiffFace\_LDM}} & \multicolumn{4}{c}{\textbf{Average}} \\ 
\cmidrule(lr){3-6}\cmidrule(lr){7-10}\cmidrule(lr){11-14}\cmidrule(lr){15-18}\cmidrule(lr){19-22}\cmidrule(lr){23-26}\cmidrule(lr){27-30}
& & AP $\uparrow$ & ACC $\uparrow$ & FPR $\downarrow$ & FNR $\downarrow$ & AP & ACC & FPR & FNR & AP & ACC & FPR & FNR & AP & ACC & FPR & FNR & AP & ACC & FPR & FNR & AP & ACC & FPR & FNR & AP & ACC & FPR & FNR  \\
\multirow{2}{*}{Explicit-based} & SBIs* & 62.7 & 69.2 & 38.5 & 23.1 & 74.5 & 65.5 & 31.1 & 38.1 & 47.4 & 48.2 & 74.3 & 28.2 & 74.6 & 69.9 & 24.8 & 35.4 & 65.2 & 58.3 & 44.0 & 45.2 & 63.0 & 58.2 & 54.1 & 28.8 & 64.6 & 61.6 & 44.4 & 33.1 \\
& CADDM* & 46.8 & 37.3 & 83.6 & 41.5 & 44.6 & 50.3 & 60.3 & 41.1 & 48.1 & 44.7 & 65.8 & 44.7 & 18.4 & 51.1 & 81.2 & 40.7 & 23.5 & 50.6 & 67.2 & 42.7 & 53.3 & 55.6 & 53.7 & 41.6 & 39.1 & 48.2 & 68.6 & 42.1\\
\cmidrule(lr){1-30}
\multirow{5}{*}{Implicit-based} & Xception & 44.6 & 50.0 & 100 & \textbf{0} & 46.6 & 50.0 & 100 & \textbf{0} & 48.5 & 50.00 & 100 & \textbf{0} & \textbf{100} & \textbf{100} & \textbf{0} & \textbf{0} & \textbf{100} & \textbf{100} & \textbf{0} & \textbf{0} & 38.0 & 50.0 & 100 & \textbf{0} & 62.9 & 66.7 & 66.7 & \textbf{0} \\
& Wang2020 & 89.3 & 55.7 & 88.6 & \textbf{0} & 76.1 & 50.8 & 98.2 & \underline{0.2} & 79.4 & 50.0 & 98.4 & \textbf{0} & 69.9 & 78.9 & 90.9 & 0.1 & 65.2 & 50.1 & 99.4 & \underline{0.4} & 54.7 & 50.0 & 100 & \textbf{0} & 72.4 & 55.9 & 95.9 & \underline{0.1} \\
& Grag2021 & \underline{99.8} & 85.7 & 28.6 & \textbf{0} & \underline{97.1} & 61.9 & 76.0 & \textbf{0} & \underline{95.2} & 55.4 & 89.2 & \textbf{0} & \textbf{100} & \underline{99.9} & \underline{0.7} & \textbf{0} & 80.9 & 79.4 & 41.2 & \textbf{0} & 53.0 & 50.0 & 100 & \textbf{0} & \underline{87.7} & 72.0 & 55.9 & \textbf{0} \\
& Ojha2023 & 88.9 & \underline{91.6} & \underline{6.6} & \underline{10.2} & 90.3 & \textbf{94.1} & \textbf{2.6} & 9.2 & 89.3 & \textbf{93.4} & \underline{3.0} & \underline{10.2} & 88.2 & 92.1 & 6.1 & 9.6 & 87.2 & \underline{92.0} & 3.6 & 12.4 & \underline{81.0} & \textbf{83.6} & \textbf{26.6} & \underline{6.3} & 87.5 & \textbf{91.1} & \underline{8.1} & 9.7\\
\cmidrule(lr){1-30}
& \textit{\textbf{NTF(Ours)}} & \textbf{99.8} & \textbf{91.8} & \textbf{0.4} & 16.0 & \textbf{99.9} & \underline{92.2} & \underline{3.3} & 15.4 & \textbf{99.9} & \underline{92.0} & \textbf{0.2} & 15.8 & \underline{99.9} & 87.8 & \textbf{0} & 15.7 & \underline{99.7} & 91.8 & \underline{0.8} & 15.6 & \textbf{84.6} & \underline{74.3} & \underline{35.6} & 15.8 & \textbf{97.3} & \underline{88.3} & \textbf{6.7} & 15.7 \\
\bottomrule
\end{tabular}
% \vspace{-10pt}
\caption{\small\textbf{Cross-family generalization on multi-step methods.} Performance of NTF and baselines in spotting 6 multi-step generated images.}
\label{tab: MultiStep}
% \vspace{-10pt}
\end{table*}

\section{Our Method}

%to achieve a universal fake image detection and avoid the classifier's reliance on specific artifact patterns,
In this work, we propose a novel method, named Natural Trace Forensics (NTF), which involves training the classifier using the natural traces shared merely by real images as an additional predictive target. Figure \ref{fig: method} overviews the pipeline of NTF. We start by learning natural trace representations from real datasets. Then, under a soft contrastive learning (SCL) framework, the network is trained to align natural traces closer to real images and further from fake ones. With such constraints, the network is motivated to detect fakes based on the distance between images and natural traces. Next, we elaborate on how to learn the natural traces and apply the extracted traces for detection.

\subsection{Natural Trace Representation Learning}

We first explore the natural traces in real images to provide learnable features for the next fake image identification. However, it is impossible to analyze every real image in existence to identify shared features. To address this, we employ an innovative strategy using the same intrinsic features found in real images as a substitute for shared features. These intrinsic features, known as homogeneous features, are derived from the inherent properties and statistical regularities of images, which are commonly present in real images. We develop a self-supervised feature mapping mechanism to extract the homogeneous features. This mechanism decouples the natural image features into homogeneous features and heterogeneous features, where the latter are associated with specific images. As opposed to direct embedding features, feature decoupling not only enhances generalization by ensuring homogeneous features more accurately capture commonalities across images, but also improves feature quality by eliminating noise and redundant information. Additionally, the network should access a large variety of natural images. This exposure enables the network to learn the patterns of feature coupling in various types of real images so that it can decouple features for unseen images. 

\noindent\textbf{Formulation.} We assume access to a large-scale dataset of real images $ D_r $. Sample $x \in D_r$ is an augmented sample from an image $x_r \in \mathbb{R}^{H \times W \times 3}$. As shown in Figure \ref{fig: method}(a), our architecture consists of a feature encoder followed by two projection heads that map real image embeddings into homogeneous and heterogeneous features. 

Specifically, feature encoder, $E$, producing feature embedding $e=E(x_r)$ from the inputs, then decouples the $e$ into homogeneous and heterogeneous features, $z^{hom} = f^{hom}(e) \in \mathbb{R}^C$ and $z^{het} = f^{het}(e) \in \mathbb{R}^C$ through two projection heads, $f^{hom}$ and $f^{het}$, respectively, where $C$ is the dimensionality of the features. Let $i\in I\equiv \{1...2N\}$ be the index of an arbitrary augmented sample, where there are a total of N samples in a batch, each with two random augmentations ($2N$). The self-supervised contrastive loss of heterogeneous (het.) feature representations in real images can be formulated as follows:
\begin{equation}
\setlength\abovedisplayskip{3pt}%shrink space
\setlength\belowdisplayskip{0.7pt}
        \mathcal{L}_{het} = -\sum_{i\in I}\log \frac{exp(z^{het}_i\cdot z^{het}_p /\tau)}{\sum_{k\in K(i)}exp(z^{het}_i\cdot z^{het}_k / \tau)}, \label{eq2:het}
\end{equation}
where the index $i$ is called the anchor, the index $p$ is called the \textit{positive}, $\tau$ is a temperature hyperparameter and $K(i)\equiv I \backslash \{i\}$, which contains all the augmented samples except $i$ in a batch. Furthermore, the loss of homogeneous (hom.) feature representations in real images can be formulated as:
\begin{equation}
\setlength\abovedisplayskip{3pt}%shrink space
\setlength\belowdisplayskip{0.7pt}
    \mathcal{L}_{hom}=\arg\max_{i\in I, k\in K'(i)} \parallel z^{hom}_k - z^{hom}_i \parallel^2_F, \label{eq1:hom}
\end{equation}
where $\parallel \cdot \parallel_F$ denotes the Frobenius norm, $K'$ contains $2N-2$ augmented samples, \ie{}, target $i$ is compared only with samples from different sources. 

Considering the potential issues of high feature coupling in the embedding space, we further use soft orthogonality (orth.) to reduce information redundancy and dependencies between these homogeneous and heterogeneous features:
\begin{equation}
\setlength\abovedisplayskip{3pt}%shrink space
\setlength\belowdisplayskip{0.7pt}
    \mathcal{L}_{orth} = \sum_{i\in I} \cos (z^{hom}_i, z^{het}_i), \label{eq3:ort}
\end{equation}

Finally, we combine these constraints to form the natural trace representation learning loss:
\begin{equation}
\setlength\abovedisplayskip{3pt}%shrink space
\setlength\belowdisplayskip{0.7pt}
    \mathcal{L}_{tra} = \mathcal{L}_{hom} + \mathcal{L}_{het} + \lambda \mathcal{L}_{orth}, \label{eq4:tra}
\end{equation}
where $\lambda$ is a scaling factor.

\subsection{Transfer Learning for Fake Image Detection}

To capture the natural traces as an additional target, we develop the SCL to simultaneously encode real and fake images further for fake image detection. Specifically, we incorporate these natural traces into SCL and constrain their distance from positive and negative samples. This will motivate the detector to identify fake images based on distance. Note that the feature encoder is frozen during this stage.

\noindent\textbf{Formulation.} We now assume access a full dataset, $D=D_r\bigcup D_f$, where $D_r$ is used in the previous stage, $D_f$ is a dataset of fake images. Our architecture consists of a feature encoder, an auxiliary projection head for SCL, and another classification head for supervised classification(shown in Figure \ref{fig: method}(b). With the auxiliary projection head, the real and fake feature embeddings are mapped to $z^r_i$ and $z^f_i$, respectively. To motivate the network to focus on the intra-class aggregation of real images more than the inter-class differences between real and fake images, we adopt the homogeneous features $z^{hom}$ from the previous stage as extra positive instances for real anchors and additional negative instances for fake anchors. SCL mitigates the negative impact of fake image features on the classifier by assigning weights to the homogeneous features. Formally, the soft contrastive loss is as follows:
\begin{equation}
\setlength\abovedisplayskip{3pt}%shrink space
\setlength\belowdisplayskip{0.7pt}
    \mathcal{L}_{scl} = \sum_{i\in I}\frac{-1}{|P(i)|}\sum_{p\in P(i)}\sum_{j\in J}\frac{exp[z_i (z_p + y\cdot\eta\cdot z^{hom}_j) /\tau]}{\sum_{k\in K(i)}exp(z_i\cdot z_k / \tau)},
\end{equation}
where $P(i)$ is the positive samples set for the anchor $i$, $|P(i)|$ is its cardinality, $K(i)$ is the negative set and $J$ is homogeneous set in a batch. Note that for real anchor, $y = 1$, otherwise $y = -1$. $\eta$ is a balance factor. To achieve fake image detection, for a given sample and label, the classifier $D$ is optimized on the binary cross entropy loss:
\begin{equation}
\setlength\abovedisplayskip{3pt}%shrink space
\setlength\belowdisplayskip{0.7pt}
    \mathcal{L}_{ce}=-\frac{1}{N}\sum_i^N y_i \cdot \log(\hat{y_i})+(1-y_i)\cdot \log(1-\hat{y_i}),
\end{equation}

Finally, the discriminative loss is given by:
\begin{equation}
    \mathcal{L}_d = \mathcal{L}_{scl}+\gamma \mathcal{L}_{ce},
\end{equation}
where $\gamma$ is a balance factor. Please refer to the technical appendix for more implementation details.

% Figure: ablation architecture
\begin{figure*}[t]
\centering
\includegraphics[width=0.9\linewidth]{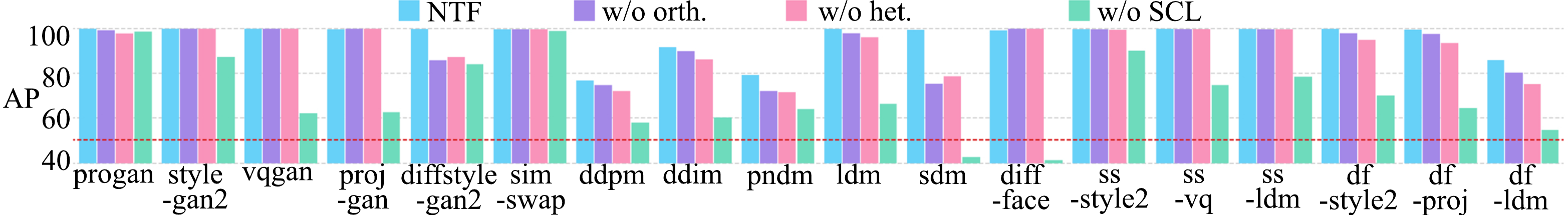}
% \vspace{-15pt}
\caption{\small\textbf{Ablation study on NTF architecture.} All detectors were trained using the ProGAN and tested on other generative models. The designs of NTF architecture improve generalization ability. The red dotted line depicts chance performance.}
\label{fig: ablation arch}
% \vspace{-10pt}
\end{figure*}

% Figure: ablation training set
\begin{figure*}[t]
\centering
\includegraphics[width=0.9\linewidth]{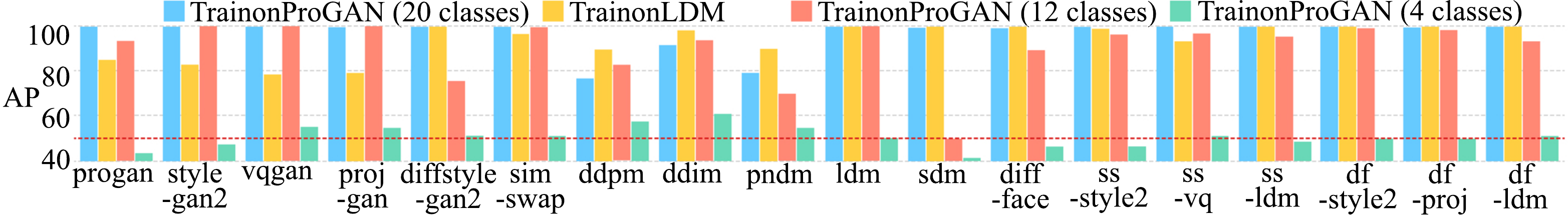}
% \vspace{-15pt}
\caption{\small\textbf{Ablation study on training data.} All detectors trained on different data sources (ProGAN or LDM) or different numbers of classes of the ProGAN data source (20 classes, 12 classes, and 4 classes).}
\label{fig: ablation data}
% \vspace{-18pt}
\end{figure*}

\begin{table}
\scriptsize
\centering
\begin{tabular}{ccccc}
    \toprule
    \textbf{Method} & Wang2020 & Grag2021 & Ojha2023 & NTF\\
    \cmidrule(lr){1-1}\cmidrule(lr){2-5}
    \textbf{Midjourney}/Acc(\%) $\uparrow$ & \underline{63.02} & 57.88 & 56.82 & \textbf{78.41}\\
    \textbf{Kolors}/Acc(\%) $\uparrow$ & 48.26 & \underline{60.24} & 57.69 & \textbf{78.12}\\
    
    \bottomrule
\end{tabular}
% \vspace{-8pt}
\caption{\small\textbf{Evaluation on commercial generative models}.}
% \vspace{-18pt}
\label{tab: commercial}
\end{table}

\section{Experiments}

\subsection{Experiments Setup}

\noindent\textbf{Dataset.} We use the dataset provided by \cite{CNNDetection}, which consists of 720K images for training and 4K images for validation. The fake images were generated by ProGAN\cite{Progan}, while an equal number of real images were sourced from LSUN\cite{lsun}. We conduct evaluation experiments on the self-built dataset, which covered GAN-based, DM-based, and multi-step fake images.

\noindent\textbf{Evaluation Metrics.} In evaluating the performance in spotting fake images synthesized with diverse generative models, we adopt four popular metrics to get a comprehensive result of our proposed method. Specifically, we report ACC (accuracy), AP (average precision), FPR (false positive rate), and FNR (false negative rate), respectively.

\noindent\textbf{Baselines.} We compare with the following six baselines, including fake detectors based on explicit and implicit artifacts: 1) Xception~\cite{rossler2019faceforensics++} is widely employed as the baseline in the studies of DeepFake forensics; 2) Wang2020\cite{CNNDetection} focuses on the artifacts exposed by CNN-generated images; 3) Grag2021\cite{gragnaniello2021gan} uses spectral super-resolution to reconstruct visual cues for detection. 4) Ojha2023\cite{UFD} uses a feature space not explicitly trained to distinguish real from fake images. 5) SBIs\cite{Self-Blended} mixes image pairs with various masks to generate training data. 6) CADDM\cite{CADDM} focuses on local information so that the network ignores identity information leakage caused by irregular face changes. Except that SBIs and CADDM are trained on FaceForensics++ \cite{rossler2019faceforensics++}, the training set for the other baselines is consistent with ours. Note that SBIs and CADDM are limited to DeepFake datasets, and for the sake of fairness, they are not included in the test of non-face data.

\noindent\textbf{Implementation Details.}
We use ResNet50 pre-trained on ImageNet as the feature encoder. All the projection heads contain two layers of MLPs with an output dimension of 128. For all datasets, we use a $224\times224$ crop for both training and testing (random crop for training and center crop for testing). For soft contrastive learning, we perform a random crop of the input image to 32 px. The optimizer is an SGD with a momentum of 0.9, an initial learning rate of 0.1, and an attenuation of 0.001. In the first stage, training is conducted for 200 epochs, followed by 10 epochs in the second stage. The empirical setting for $\lambda$, $\eta$, $\gamma$ is set at 0.1, 0.5, 0.5.

% Figure: robust
\begin{figure*}[t]
\centering
\includegraphics[width=\linewidth]{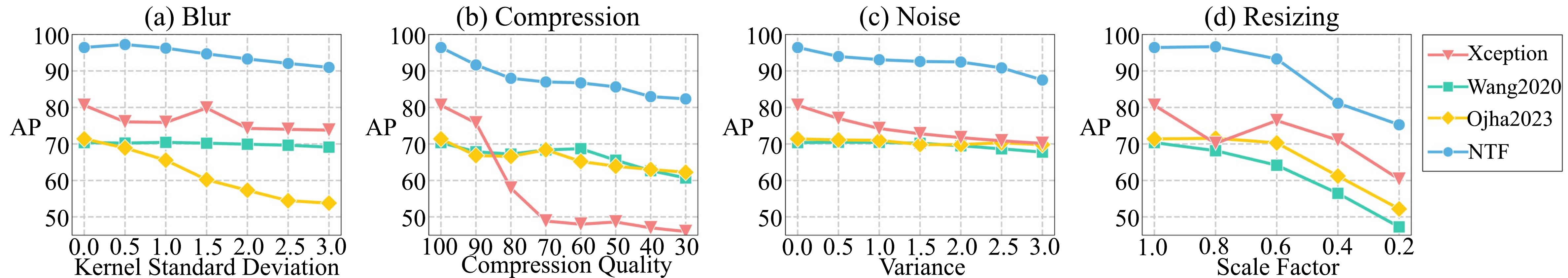}
% \vspace{-15pt}
\caption{\small Robustness to four image processing operations, \ie{}, Gaussian blur (a), JPEG compression (b), Gaussian Noise (c), Scaling (d).}
% \vspace{-10pt}
\label{fig: robust}
\end{figure*}

% Figure: Heatmap
\begin{figure*}[t]
\centering
\includegraphics[width=0.9\linewidth]{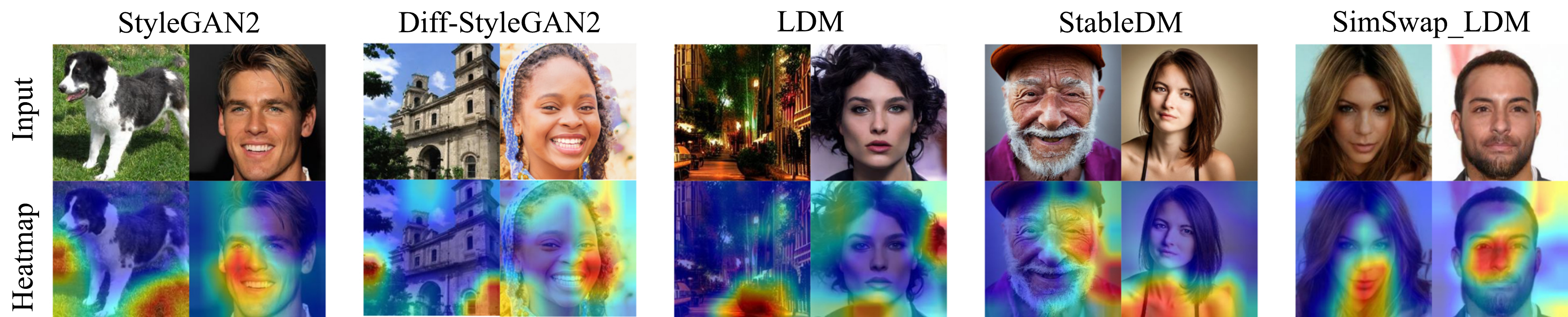}
% \vspace{-5pt}
\caption{\small The Grad-CAM++ visualization of the self-built dataset.}
\label{fig: heatmap}
% \vspace{-15pt}
\end{figure*}

\subsection{Effectiveness Evaluation}
We explore the capability of NTF to detect fake images from different generative models in the self-built dataset.

\noindent\textbf{Performance on GAN-based fake images.} 
The results in Table \ref{tab: GANbased} demonstrate the capability of NTF in tackling unknown GAN-based fake images with the highest AP \textbf{99.9\%} and the lowest FPR \textbf{0.2\%} on average. In particular, NTF attains optimal performance on GAN-based fake images, which could be attributed to the similarity in the image generation principle of these models to that of ProGAN. That is precisely why certain baselines, such as Xception and Grag2021, have achieved high-performance generalization on GAN-based generative models. This indicates that existing fake image detections are adept at handling generalization scenarios within the same model family.

\noindent\textbf{Performance on DM-based fake images.}
As illustrated in Table \ref{tab: DMbased}, NTF exhibits superior cross-family generalization capability on the DMs with the highest AP \textbf{91.2\%} and ACC \textbf{82.7\%} on average. Specifically, NTF improves the AP and ACC by nearly \textbf{6\%} and \textbf{19.2\%} compared to the best baseline. The detection accuracy of NTF surpasses all baselines across four DMs. However, it remains comparable to Grag2021 on LDM and to XeceptionNet on DDPM. Overall, NTF detects different DM-base generation models in a more balanced way, although it is slightly inferior to some baselines on certain models.

\noindent\textbf{Performance on multi-step fake images.}
As shown in Table \ref{tab: MultiStep}, NTF detects fake images synthesized by six multi-step methods with the highest mAP of \textbf{97.3\%}. It is noteworthy that the method from Ojha2023 demonstrated an average ACC of 91.1\%, slightly outperforming NTF. This can be attributed to the use of ViT-L/14\cite{ViT}, a vision transformer variant pre-trained on CLIP\cite{CLIP} for the backbone, aiding in effectively modeling details for real and fake image classification.

\noindent\textbf{Performance on commercial generative models.} To better evaluate the performance in tackling commercial generative models, we assess the detection capability of NTF on fake images generated by \textit{Midjourney} and \textit{Kolors} in Table \ref{tab: commercial}. Experimental results show that our proposed method NTF gives an accuracy more than \textbf{78.4\%} which significantly outperforms the baselines. 

In summary, NTF effectively detects GAN-based unknown generative models with a mAP \textbf{99.9\%}. It also demonstrates the capability to detect DM-based and multi-step generation methods with mAPs of \textbf{91.2\%} and \textbf{97.3\%} respectively, achieving an overall mAP of \textbf{96.2\%} across all datasets. Additionally, NTF shows a slightly higher FNR, which can be attributed to the inadequate coverage of real-world images in the training dataset.

\subsection{Ablation Studies}

We evaluated the effect of architecture and training data on the generalization ability of NTF. 
It initially employed homogeneous, heterogeneous, and soft orthogonality losses with Eq (\ref{eq4:tra}) for learning natural trace representations and it trains with ProGAN/LSUN dataset consisting of 20 classes. 

\noindent\textbf{Effect of network architecture.} 
We conducted experiments with different variants of NTF, exploring the following configurations: 1) without \textit{ort.} loss, 2) without \textit{het.} and \textit{orth.} loss and 3) without SCL. For each, We maintain ProGAN real/fake image data as training data. Figure \ref{fig: ablation arch} shows the performance of these variants on the same models. We find that the architectural design of NTF has an important role in generalization, improving performance on conditional generative models such as LDM, SDM, and multi-step synthesis methods. This improvement is likely attributed to the implementation of soft orthogonality and heterogeneous loss constraints on upper bounds, which enables the NTF to learn homogeneous features capable of effectively separating traces of conditional generative models from natural traces. Moreover, this design also improves performance on unconditional generative models, particularly more significant in models like Diff-StyleGAN and PNDM.

\noindent\textbf{Effect of training data}
Next, we investigate how the source and diversity of the training data influence a detector's generalization ability. Our approach involves altering either the data source or the number of classes in the training set. Specifically, we train multiple detectors by 1) employing a pre-trained LDM, substituting the ProGAN, and 2) utilizing a subset of the full ProGAN dataset, excluding real and fake images from certain LSUN classes, as shown in Figure \ref{fig: ablation data}. With access only to the LDM dataset, the model displays impressive generalization capabilities, even though the dataset consists of 200k reals from LAION\cite{LAION} and 200k fakes generated by LDM. As expected, diversifying the training set does enhance generalization to some extent, but the benefits diminish with increasing diversity. This suggests the potential existence of a real image dataset capable of extracting universally present natural traces.

\subsection{Robustness Evaluation}
In this section, we mainly explore the robustness of our proposed method in surviving diverse input transformations, such as blur, compression, noise, and scaling. 
%\wang{Note that our NTF, Wang2020, and Ojha2023 train with jpeg and blur data augmentation on ProGAN real and fakes.}

As shown in Figure \ref{fig: robust}, NTF is typically robust to blur, compression, and noise operations. The performance of Wang2020 is more consistent under different levels of blur/compression/noise. This is reasonable, as it employs random compression and blur for data augmentation during training. Although Ojha2023 also employs data augmentation, it also performs in addition to being less robust in blur operations. In addition, both NTF and other baselines show significant performance degradation in image scaling operations. When the image scaling factor is 0.2, the image content becomes imperceptible, making robustness in such extreme scenarios less critical. Please refer to the technical appendix for more details.

\subsection{Qualitative Analysis}
To understand how the network generalizes to different synthesis methods, we visualize the model saliency map.
We apply GradCAM++ \cite{Grad-CAM} to NTF on the self-built dataset to visualize where models are paying their attention to images, as shown in Figure \ref{fig: heatmap}. This demonstrates how the network captures artifacts from different generative models; for example, with the dog generated by StyleGAN2, the network focuses on the inconsistency in shadow.
% \vspace{-5pt}

\section{Conclusion}
In this paper, we escape the trap of exploring subtle differences between real and fake for fake image detection. Motivated by the presence of common shared features in real images, We propose a novel framework named NTF, pre-trained by natural trace representation learning and soft contrastive learning, to significantly improve the generalization ability of fake image detection. Extensive experiments on the self-built dataset demonstrate that our method exhibits state-of-the-art generalization capability for unknown generative models. Our research also offers a fresh perspective on fake image detection,  focusing on exploring stable detectable features rather than those that continuously change, thereby paving the way for future studies in this field.

% \section{Technical Appendix}
% In the technical appendix, we present the details of the self-built dataset, evaluation of the test set from \cite{CNNDetection}, some visualizations, and additional experiments. 

\section{Acknowledgements}
This research was supported in part by the National Key Research and Development Program of China under No.2021YFB3100700, the National Natural Science Foundation of China (NSFC) under Grants No. 62202340, 62372334, the CCF-NSFOCUS `Kunpeng' Research Fund under No. CCF-NSFOCUS 2023005, the Open Foundation of Henan Key Laboratory of Cyberspace Situation Awareness under No. HNTS2022004, the Fundamental Research Funds for the Central Universities under No. 2042023kf0121.

\bibliography{aaai25}

\end{document}

\maketitle

% Complementary Robustness
\begin{figure}[t]
    \centering
    \includegraphics[width=1\linewidth]{img/pdf/sup-robust_compressed.pdf}
    \caption{Robustness evaluation of fake image detection methods (NTF and three baselines) under four types of image perturbations (Gaussian blur, JPEG compression, Gaussian noise, and image resizing). Each row represents the robustness evaluation results of fake image detection methods on GAN models, Diffusion models, and Multi-step models, respectively. These results reveal the robustness of the NTF method in detecting fake images under various types of image disturbances.}
    \label{fig:sup-robust}
\end{figure}

\begin{abstract}
In this supplementary material, we present extensive experiments to explore the effectiveness of our proposed method in tackling some public datasets and provide a complementary evaluation of robustness. Moreover, we also provide details of our dataset collection.

\begin{itemize}[leftmargin=*]
\item We present additional experiment results and qualitative analysis for generalization evaluation.
\item We show the philosophy of dataset collection.
\end{itemize}
\end{abstract}

% public dataset
\begin{table*}[t]
\scriptsize
\centering
\setlength{\tabcolsep}{3pt}
\begin{tabular}{ccccccccccccc}
\toprule
\multirow{2}{*}{\textbf{Method}} & \multicolumn{3}{c}{Unconditional GAN} & \multicolumn{3}{c}{Conditional GAN} & \multicolumn{2}{c}{Perceptual loss} & \multicolumn{2}{c}{Low-level vision} & DeepFake & Average \\
\cmidrule(lr){2-4}\cmidrule(lr){5-7}\cmidrule(lr){8-9}\cmidrule(lr){10-11}\cmidrule(lr){12-12}
& ProGAN & StyleGAN & BigGAN & CycleGAN & StarGAN & GauGAN & CRN & IMLE & SITD & SAN & FF++ & \textbf{mAP} \\ 
\midrule
XceptioNet \cite{rossler2019faceforensics++} & \textbf{100} & 94.31 & 66.96 & 90.55 & \textbf{100} & 60.91 & \textbf{96.35} & 56.83 & \underline{64.49} & 58.83 & 58.78 & 67.95 \\
Wang2020 \cite{CNNDetection} & \textbf{100} & 96.82	& 88.23	& \underline{98.51}	& \underline{95.46} & \underline{98.09}	& 66.27	& \textbf{92.72}	& 63.88	& \underline{98.95}	& \underline{99.52} & \underline{96.86} \\
Grag2021 \cite{gragnaniello2021gan} & 99.16	&\textbf{99.01}	&72.80	&84.95	&\textbf{100}	&66.47	&\underline{96.29}	&58.15	&51.80	&59.76	&89.54	&79.57\\
Ojha2023 \cite{UFD} & \underline{99.70}	&97.67	&\underline{94.21}	&80.25	&92.84	&\textbf{99.10}	&66.49	&69.44	&51.64	&56.57	&69.09	&79.43 \\
NTF [Ours] & \textbf{100}	&\underline{98.45}	&\textbf{94.37}	&\textbf{100}	&\textbf{100}	&92.77	&73.58	&\underline{81.89}	&\textbf{87.94}	&\textbf{99.99}	&\textbf{100}	&\textbf{97.16}\\

\bottomrule
\end{tabular}

\caption{\textbf{Evaluations of various fake image detection methods on public datasets}. The performance of each method is measured by Average Precision (AP) scores, with the best value being indicated in \textbf{bold} and the second-best value \underline{underlined}. The \textit{Average} column represents the mean AP scores across all datasets.}

\label{tab:Public}
\end{table*}

\section{Extensive Experimental Results} \label{sec:more}
\subsection{Complementary Robustness Evaluation}
We present additional robustness evaluations of fake image detection methods that suffer from various image perturbations, as shown in Figure \ref{fig:sup-robust}. Specifically, we demonstrate the robustness evaluation of fake detections on GANs, diffusion models, and multi-step models, subjected to Gaussian blur, JPEG compression, Gaussian noise, and image scaling perturbations. We observe that our proposed method, NTF, exhibits robustness against blur, compression, and noise. Additionally, NTF demonstrates similar capabilities across different families, indicating a robust trend on other unknown generative models as well.

\subsection{Evaluation on the Public Datasets}
To assess the generalization ability of our method beyond overfitting to our custom dataset, we also evaluate its capabilities on other publicly available datasets. Specifically, we conduct evaluations on the test dataset released by \cite{CNNDetection}, as shown in Table \ref{tab:Public}. The dataset consists of ProGAN~\cite{Progan}, StyleGAN~\cite{Stylegan}, BigGAN~\cite{Biggan}, CycleGAN~\cite{Cyclegan}, StarGAN~\cite{Stargan}, GauGAN~\cite{Gaugan}, CRN~\cite{CRN}, IMLE~\cite{IMLE}, SAN~\cite{SAN}, SITD~\cite{SITD}, and DeepFakes~\cite{rossler2019faceforensics++}, each comprising collections of real and fake images. 

We observe that NTF exhibits significant performance across both unconditional GANs and conditional GANs. This is reasonable since NTF is trained on ProGAN and has already demonstrated intra-family generalization on our self-built dataset. However, NTF does not achieve the best performance on GauGAN, which may be attributed to the inclusion of a substantial number of resized images in that dataset. Notably, NTF demonstrates remarkable generalization ability on low-level vision and DeepFakes, while its performance on the perceptual loss metric is suboptimal. We attribute this to the fact that the datasets used by CRN and SAN methods are sourced from GTA, a dataset consisting of video game visuals. NTF easily identified supposedly real game visuals as fake or unnatural due to the discrepancy in dataset sources.

\subsection{Additional visualization}
\begin{figure}[t]
\centering
\includegraphics[width=\linewidth]{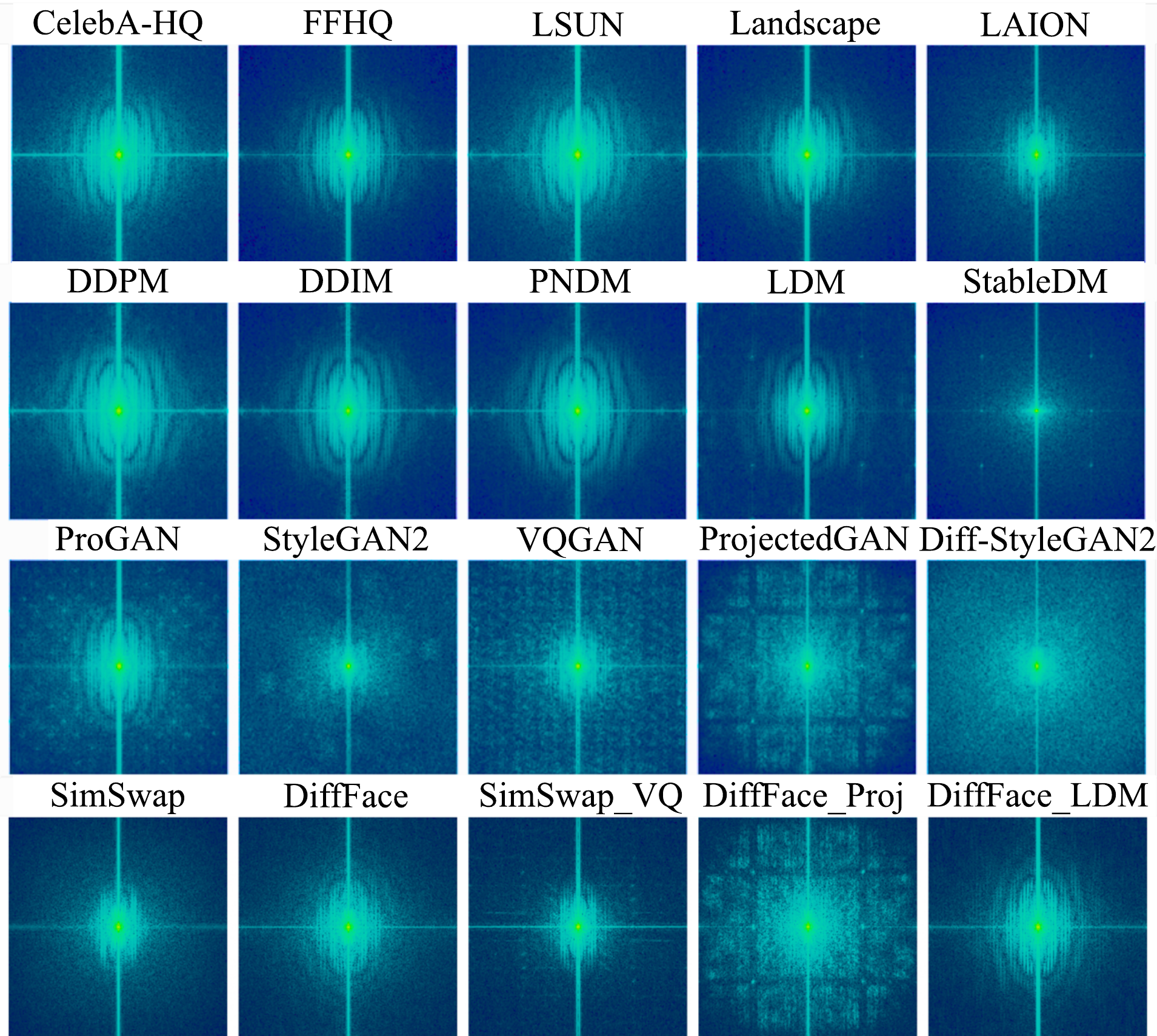}

\caption{Frequency analysis on the self-built dataset. We show the averaged spectra of each model. }
\label{fig:freartifact}

\end{figure}

We visualize the average frequency spectra from the self-built dataset to study the artifacts from different generative models, as shown in Figure \ref{fig:freartifact}. Although the real image spectra look similar, there are obvious patterns in the images synthesized with different generative models. What's more, there are significant differences in the patterns of different generative models, suggesting that the artifact of only learning a single model is a trap that leads to poor generalization of existing work.

% Figure: freartifact

\section{Self-built Dataset Collection} \label{sec:col}
We collect available pre-trained models from publicly available code repositories or directly download the available generated images. To ensure the quality of the generated images, we select the pre-trained models with the highest FID scores and follow the recommendations of each method. Then we pre-processed the real images (\eg{}, resize, and crop) according to each method's pipeline to provide the real distribution of real and false images as close as possible.

\noindent\textbf{ProGAN} \cite{Progan} We take the officially released ProGAN model pre-trained on CelebA-HQ~\cite{Progan} with size $1024\times1024$. We download the released synthesized images, and following the code, we download the real images from Kaggle by resizing both real and fake to $256\times256$.

\noindent\textbf{StyleGAN2} \cite{Styleganv2} We take officially released StyleGAN2 models pre-trained on FFHQ\cite{Stylegan}, CelebA-HQ and LSUN\cite{lsun} dog with size $1024\times1024$, $256\times256$ and $256\times256$. We generate fake images with pre-trained models following the code and download the real images from each category.

\noindent\textbf{ProjectedGAN} \cite{Projectedgan} We take officially released ProjectedGAN models pre-trained on FFHQ, LSUN bedroom and Landscape\footnote{\noindent https://www.kaggle.com/datasets/arnaud58/landscape-pictures} with $256\times256$. We generate fake images with pre-trained models following the code and download the real images from each category.

\noindent\textbf{VQGAN} \cite{Vqgan} We take the officially released VQGAN model pre-trained on CelebA-HQ with size $256\times256$. We directly download the released real and fake images.

\noindent\textbf{Diff-StyleGAN2} \cite{Diffusiongan} We take officially released Diff-StyleGAN2 models pre-trained on FFHQ, LSUN bedroom and church with size $1024\times1024$, $256\times256$ and $256\times256$. We generate fake images with pre-trained models following the code and download the real images from each category.

\noindent\textbf{SimSwap} \cite{chen2020simswap} We take the officially released SimSwap pre-trained model on VGGFace2-HQ with size $224\times224$. We perform face-swapping on CelebA-HQ to generate fake images and collect the corresponding background faces as real images. 

\noindent\textbf{DDPM} \cite{DDPM} We take the officially released DDPM pre-trained model on CelebA-HQ with $256\times256$.  We generate fake images with pre-trained models following the code and download the real images.

\noindent\textbf{DDIM} \cite{DDIM} We take the officially released DDIM pre-trained model on CelebA-HQ with $256\times256$. We generate fake images with pre-trained models following the code and download the real images.

\noindent\textbf{PNDM} \cite{PNDM} We take the officially released PNDM pre-trained model on CelebA-HQ with $256\times256$. We generate fake images with pre-trained models following the code and download the real images.

\noindent\textbf{LDM} \cite{LDM} We take the officially released LDM pre-trained model on CelebA-HQ and LAION-400M\footnote{https://laion.ai/blog/laion-400-open-dataset} with $256\times256$ and $256\times256$. We generate fake images with pre-trained models following the code and download the corresponding real images.

\noindent\textbf{StableDM} \cite{LDM} We take the officially released StableDM pre-trained model on LAION-2B-en\footnote{https://huggingface.co/datasets/laion/laion2B-en} with size $512\times512$. We generate fake images with pre-trained models following the code and download real images with corresponding prompts.

\noindent\textbf{DiffFace} \cite{kim2022diffface} We take the officially released DiffFace pre-trained model on FFHQ with size $256\times256$. We perform face-swapping on CelebA-HQ to generate fake images and collect the corresponding background faces as real images.

Moreover, we collect real faces as backgrounds and swap them with other generated faces for multi-step methods.

\section{Reproducibility Checklist}

This paper:

\begin{itemize}[leftmargin=*]
\item Includes a conceptual outline and/or pseudocode description of AI methods introduced (yes)
\item Clearly delineates statements that are opinions, hypothesis, and speculation from objective facts and results (yes)
\item Provides well marked pedagogical references for less-familiare readers to gain background necessary to replicate the paper (yes)
\end{itemize}

\noindent Does this paper make theoretical contributions? (yes)

\noindent If (yes), please complete the list below.

\begin{itemize}[leftmargin=*]
\item All assumptions and restrictions are stated clearly and formally. (yes)
\item All novel claims are stated formally (e.g., in theorem statements). (yes)
\item Proofs of all novel claims are included. (yes)
\item Proof sketches or intuitions are given for complex and/or novel results. (yes)
\item Appropriate citations to theoretical tools used are given. (yes)
\item All theoretical claims are demonstrated empirically to hold. (yes)
\item All experimental code used to eliminate or disprove claims is included. (yes)
\end{itemize}

\noindent Does this paper rely on one or more datasets? (yes)

\noindent If (yes), please complete the list below.

\begin{itemize}[leftmargin=*]
\item A motivation is given for why the experiments are conducted on the selected datasets (yes)
\item All novel datasets introduced in this paper are included in a data appendix. (yes)
\item All novel datasets introduced in this paper will be made publicly available upon publication of the paper with a license that allows free usage for research purposes. (yes)
\item All datasets drawn from the existing literature (potentially including authors’ own previously published work) are accompanied by appropriate citations. (yes)
\item All datasets drawn from the existing literature (potentially including authors’ own previously published work) are publicly available. (yes)
\item All datasets that are not publicly available are described in detail, with explanation why publicly available alternatives are not scientifically satisficing. (yes)
\end{itemize}

\noindent Does this paper include computational experiments? (yes)

\noindent If (yes), please complete the list below.

\begin{itemize}[leftmargin=*]
\item Any code required for pre-processing data is included in the appendix. (yes)
\item All source code required for conducting and analyzing the experiments is included in a code appendix. (yes)
\item All source code required for conducting and analyzing the experiments will be made publicly available upon publication of the paper with a license that allows free usage for research purposes. (yes)
\item All source code implementing new methods have comments detailing the implementation, with references to the paper where each step comes from (yes)
\item If an algorithm depends on randomness, then the method used for setting seeds is described in a way sufficient to allow replication of results. (yes)
\item This paper specifies the computing infrastructure used for running experiments (hardware and software), including GPU/CPU models; amount of memory; operating system; names and versions of relevant software libraries and frameworks. (yes)
\item This paper formally describes evaluation metrics used and explains the motivation for choosing these metrics. (yes)
\item This paper states the number of algorithm runs used to compute each reported result. (yes)
\item Analysis of experiments goes beyond single-dimensional summaries of performance (e.g., average; median) to include measures of variation, confidence, or other distributional information. (yes)
\item The significance of any improvement or decrease in performance is judged using appropriate statistical tests (e.g., Wilcoxon signed-rank). (yes)
\item This paper lists all final (hyper-)parameters used for each model/algorithm in the paper’s experiments. (yes)
\item This paper states the number and range of values tried per (hyper-) parameter during development of the paper, along with the criterion used for selecting the final parameter setting. (yes)
\end{itemize}

\bibliography{aaai25}